\documentclass[journal]{IEEEtran}
\usepackage{amsmath,amsfonts}
\usepackage{algorithmic}
\usepackage{array}
\usepackage[caption=false,font=normalsize,labelfont=sf,textfont=sf]{subfig}
\usepackage{textcomp}
\usepackage{stfloats}
\usepackage{url}
\usepackage{verbatim}
\usepackage{graphicx}
\def\BibTeX{{\rm B\kern-.05em{\sc i\kern-.025em b}\kern-.08em
T\kern-.1667em\lower.7ex\hbox{E}\kern-.125emX}}

\usepackage{inconsolata}
\usepackage{xcolor}
\usepackage{color}
\usepackage[colorlinks,
            linkcolor=black, 
            citecolor=black, 
            urlcolor=black, 
          ]{hyperref}
\usepackage{booktabs}
\usepackage{makecell}
\usepackage{multirow}
\usepackage{float}
\usepackage{wrapfig}
\usepackage{amssymb}
\usepackage{enumitem}

\usepackage[numbers]{natbib} 
\usepackage[T1]{fontenc}
\usepackage{orcidlink}
\usepackage{threeparttable}
\usepackage{capt-of}

\setcitestyle{numbers,open={\textcolor{black}{[}},close={\textcolor{black}{]}}}

\begin{document}
\title{AutoHall: Automated Factuality Hallucination Dataset Generation for Large Language Models}

\author{Zouying Cao$^{*\orcidlink{0000-0003-2200-5630}}$, Yifei Yang$^{*\orcidlink{0000-0003-0997-9422}}$, Xiaojing Li$^{\orcidlink{0000-0002-8473-5594}}$, Hai Zhao$^{\orcidlink{0000-0001-7290-0487}}$\thanks{Zouying Cao, Yifei Yang and Hai Zhao are with the AGI Institute, School of Computer Science, Shanghai Jiao Tong University, Shanghai 200240, China, and also with the Shanghai Key Laboratory of Trusted Data Circulation and Governance in Web3, Shanghai 200240, China (e-mail: zouyingcao@sjtu.edu.cn; yifeiyang@sjtu.edu.cn; zhaohai@cs.sjtu.edu.cn). Xiaojing Li is with the School of Media \& Communication, Shanghai Jiao Tong University, Shanghai 200240, China (e-mail: lixiaojing@sjtu.edu.cn). * denotes equal contribution.}}



\maketitle

\begin{abstract}
Large language models (LLMs) have gained broad applications across various domains but still struggle with hallucinations. 
Currently, hallucinations occur frequently in the generation of factual content and pose a great challenge to trustworthy LLMs. 
However, hallucination detection is hindered by the laborious and expensive manual annotation of hallucinatory content. 
Meanwhile, as different LLMs exhibit distinct types and rates of hallucination, the collection of hallucination datasets is inherently model-specific, which also increases the cost. 
To address this issue, this paper proposes a method called \textbf{AutoHall} for \underline{Auto}matically constructing model-specific \underline{Hall}ucination datasets based on existing fact-checking datasets. 
The empirical results reveal variations in hallucination proportions and types among different models. 
Moreover, we introduce a zero-resource and black-box hallucination detection method based on self-contradiction to recognize the hallucination in our constructed dataset, achieving superior detection performance compared to baselines. 
Further analysis on our dataset provides insight into factors that may contribute to LLM hallucinations. 
Our codes and datasets are publicly available at \url{https://github.com/zouyingcao/AutoHall}.
\end{abstract}


\section{Introduction}
\begin{figure}[t]
\centering
\includegraphics[width=0.9\linewidth]{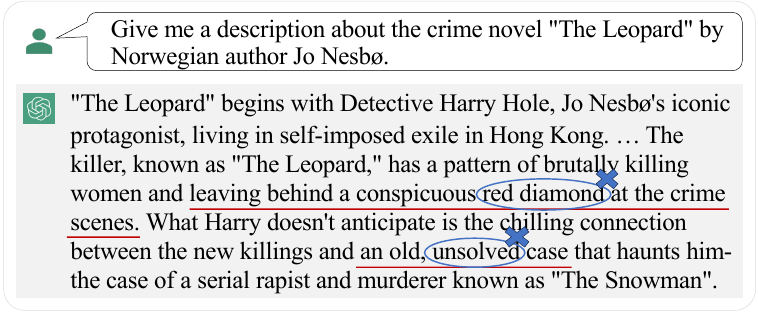}
\caption{A hallucination example. The red underline indicates the hallucinatory content generated by ChatGPT, since the novel never mentions the presence of a ``red diamond'' at the crime scene and the ``The Snowman'' case has also been solved before. }
\label{fig:hallucination}
\end{figure}

\IEEEPARstart{L}{arge} language models (LLMs) 
are capable of performing a wide range of tasks across diverse domains~\cite{sohail2023decoding,sallam2023chatgpt,chatgptapplication}. 
Despite their powerful capabilities, LLMs suffer from the issue of \textit{hallucination}, \textit{i.e.}, have the tendency to respond inaccurate or fabricated information in generation tasks~\cite{zhang2023siren,ji2023survey,rawte2023survey}. 
As shown in Fig.~\ref{fig:hallucination}, ChatGPT generates hallucinations when describing Jo Nesbø's novel ``The Leopard''. 
It can be observed that ChatGPT fabricates some plots of the novel and contains incorrect texts in the response. 
With the increasing reliance of current artificial intelligence (AI) systems on LLMs, the generation of hallucinatory information poses huge challenge to enterprise security and user trust~\cite{snowball,threatgpt}. 
Therefore, significant endeavors have been undertaken to detect LLM hallucinations. 

Current research efforts on hallucination detection leverage external knowledge bases~\cite{chern2023factool,gou2023critic,li2023self} or just adopt a zero-resource approach which focuses on resources inherent to the model itself~\cite{azaria2023internal,agrawal2023language,varshney2023stitch,selfcheckgpt,mundler2023self}.
Typically, most of these methods begin with a crowdsourced annotation, where researchers use question answering (QA) datasets to collect the model-generated responses and then manually annotate whether the answers contain hallucinations. 
However, this collection pipeline of hallucination detection datasets has several limitations. 
First, hallucination datasets are model-specific because different LLMs have varying hallucination properties. 
Even when given identical inputs, some models may generate hallucinatory content while others produce factually accurate responses.
Therefore, each model requires dedicated annotation, which limits the dataset scalability due to the dependence on manual annotation~\cite{hu2024refchecker}. 
Second, collected datasets are time-sensitive since model upgrades may mitigate hallucination issues in LLMs~\cite{cheng2023evaluating} and the old dataset can no longer represent the new model hallucinations. 

Considering the above issues, this paper explores one automated generation pipeline of hallucination detection datasets. 
Our focus is on \textit{factuality hallucination}, a primary LLM hallucination type, with studies showing GPT-4's 28\% factual hallucination rate is over double that of other types~\cite{bilan2025llm}.
Inspired by Agrawal et al.~\cite{agrawal2023language} emphasizing the hallucinatory reference\footnote{The "reference" term means the supporting information generated by the LLM to substantiate its claim, rather than a traditional bibliographic citation.} problem in LLMs, we find the feasibility of automatically creating hallucination detection datasets through public fact-checking datasets. 
Specifically, since the existing fact-checking datasets usually consist of manually annotated claims accompanied by the ground truth labels (i.e.,  factual/non-factual), we can determine whether hallucinations exist by generating references to the claims and exploring whether the references can infer the correct labels for the claims.

\IEEEpubidadjcol
Our analysis reveals that LLMs are particularly susceptible to hallucinations in responses involving various domain-specific topics such as history, technology and geography. We further investigate the contributing factors across different models based on our dataset. 
Additionally, we propose a three-step zero-resource black-box hallucination detection method motivated by the idea of self-contradiction~\cite{selfcheckgpt}. 
Given that one LLM accurately understands a claim, the generated references 
are less likely to contain contradictions. 
Therefore, it is possible to determine whether the model has generated hallucinations based on knowledge conflicts among these references. 

In summary, the contributions of our paper are: 
\begin{itemize}
\item We propose an approach called \textbf{AutoHall} for fast and \underline{auto}matically constructing model-specific \underline{hall}ucination datasets based on existing fact-checking datasets, eliminating the need for manual annotation.
\item Using AutoHall, we conduct extensive experiments to analyze LLM hallucinations in current open-source and closed-source models. From the results, we estimate the prevalence of hallucination in LLMs at a rate of 20\% to 30\% and gain insight into what types or topics of LLM responses that tend to be hallucinatory. 
\item We introduce a black-box hallucination detection method without external resources. Based on our dataset, we evaluate its effectiveness on ChatGPT and Llama2 models, demonstrating its superior improvements over existing zero-resource detection techniques.
\end{itemize}

\section{Background and Related Works}\label{related_works}
\subsection{Hallucination of Large Language Models}
Although large language models have demonstrated remarkable capabilities~\citep{liu2023summary,srivastava2022beyond}, they still struggle with several issues, where hallucination is a significant problem. 
The hallucination issue refers to that LLMs generate the statements which appear plausible but are fabricated or contradict factual knowledge. 
The consequent effects may undermine the reliability of LLM applications~\citep{zhang2023siren,pan2023risk}, thereby causing negative user experiences.

Generally, hallucinations of large language models can be divided into two primary categories: intrinsic hallucinations and extrinsic hallucinations~\citep{ji2023survey}. 
Intrinsic hallucinations occur when the LLM output contradicts the input content. 
For example, in a multi-modal image captioning task, the model generates a caption that includes details or objects which are not present in the input image. 
Extrinsic hallucinations refer to the generated content that cannot be verified based on the source content. 
Huang et al.~\cite{huang2023survey} redefine the taxonomy of hallucination by introducing factuality hallucination and faithfulness hallucination, taking alignment with user directives and factual knowledge into account. 
Another recent paper~\cite{rawte2023troubling} meticulously classify hallucination into six types according to the specific hallucinatory content.
In this paper, our focus is on \textit{factuality hallucination}. 

So far, the causes of hallucination in LLMs have been investigated across different tasks, such as question answering~\cite{zheng2023does}, abstractive summarization~\cite{cao2021hallucinated}, machine translation~\cite{guerreiro2023hallucinations} and dialogue systems~\cite{das2023diving}. 
The key factors include but are not limited to training corpora quality~\cite{mckenna2023sources,dziri2022origin}, problematic alignment process~\cite{zhang2023siren,radhakrishnan2023question} and randomness in generation strategy~\cite{lee2022factuality,dziri2021neural}. 
McKenna et al.~\cite{mckenna2023sources} offer one LLM bias brought by training text as explanation of general false positive hallucination. 
Alignment tax is also a well-known problem that may lead to LLM hallucinations~\cite{ji2024llm}. 
During inference, the existence of a likelihood trap~\cite{zhang2020trading} supports incorporating randomness into decoding strategies while exacerbates the risk of hallucinations~\cite{aksitov2023characterizing}.

\subsection{Hallucination Evaluation Datasets}
For hallucination detection benchmarks, the majority of previous studies have focused on task-specific hallucinations to support detection tasks in numerous scenarios~\cite{li2023halueval,umapathi2023med,dale2023halomi}. 
For example, Umapathi et al.\cite{umapathi2023med} propose a hallucination benchmark within the medical domain as a tool for hallucination evaluation and mitigation. 
Dale et al.~\cite{dale2023halomi} present another dataset with human-annotated hallucinations in machine translation to promote the research on translation pathology detection and analysis. 

Beyond single domain task, critically assessing the effectiveness of hallucination detection strategies necessitates the development of high-quality datasets across multiple domain tasks. 
SelfCheckGPT-Wikibio~\cite{selfcheckgpt} makes a contribution as a sentence-level hallucination detection dataset and HaluEval~\cite{li2023halueval} dataset is constructed using a combination of synthetically and naturally generated LLM responses. 
Besides, FELM~\cite{zhao2024felm} benchmark provides fine-grained factuality labels to responses generated from ChatGPT. 
Hu et al.~\cite{hu2024refchecker} curate a comprehensive dataset that enables the evaluation of hallucination detection performance under different context quality and availability. 

Nevertheless, there are limitations as they are subject to manually annotated hallucination datasets, which are expensive and time-consuming. 
Meanwhile, hallucination datasets are model-specific, requiring separate annotations for different models, whose applicability would also be affected by model upgrades (\textit{i.e.}, time-sensitive). 
Thus, there is still room for improvement in the collection process of current hallucination detection datasets.

\subsection{LLM Hallucination Detection}
To detect the hallucination issue, researchers have been making tremendous effort in seeking solutions. 
Existing approaches can be broadly categorized into two types: retrieval-based detection and zero-resource detection. 

On the one hand, prior works focus on resorting to external knowledge bases to detect hallucinations. 
For instance, Gou et al.~\cite{gou2023critic} propose a framework called CRITIC to validate the output generated by the model with tool-interaction and Huo et al.~\cite{huo2023retrieving} improve the conventional method of retrieving supporting evidence for hallucination detection by incorporating the LLM-generated answer into retrieval query. 
In addition, a unified framework called FACTOOL~\cite{chern2023factool} invokes interfaces of search engines to recognize hallucination. 

On the other hand, current researches pay more attention to realizing zero-resource hallucination detection methods. 
These methods primarily depend on uncertainty of the LLMs~\cite{chen2024quantifying}, which is mainly reflected in self-contradiction phenomenon among their responses or the model’s internal states. 
Typically, Xue et al.~\cite{xue2023rcot} utilize the Chain of Thoughts (CoT) to check the hallucinatory responses. 
Manakul et al.~\cite{selfcheckgpt} introduce a simple sampling-based approach called SelfCheckGPT that can be used to detect hallucination. 
They examine five variants of SelfCheckGPT for measuring consistency via: BERTScore, question-answering, n-gram, NLI and LLM prompting. 
With respect to open-source LLMs, the internal states of LLMs can be used to indicate the existence of hallucination, through the analysis of metrics like token probability or entropy~\cite{azaria2023internal,luo2023zero}. 

Our proposed detection method falls into the zero-resource category which is suitable for both open-source and closed-source models. 
Using AutoHall, our up-to-date constructed datasets can evaluate whether these detection methods remain consistently applicable regardless of model upgrades and then we expect our work to lay a solid foundation for subsequent research in hallucination detection. 

\section{Methodology}
In this section, we first formulate the definition of LLM factuality hallucinations discussed in our work. 
Then, we introduce AutoHall, our automatic dataset creation pipeline which focuses on prompting LLMs to produce ``hallucinatory references''. 
Finally, based on our generated datasets, we further present one zero-resource, black-box approach to recognize hallucination. 

\begin{figure*}[ht]
\centering
\includegraphics[width=1\linewidth]{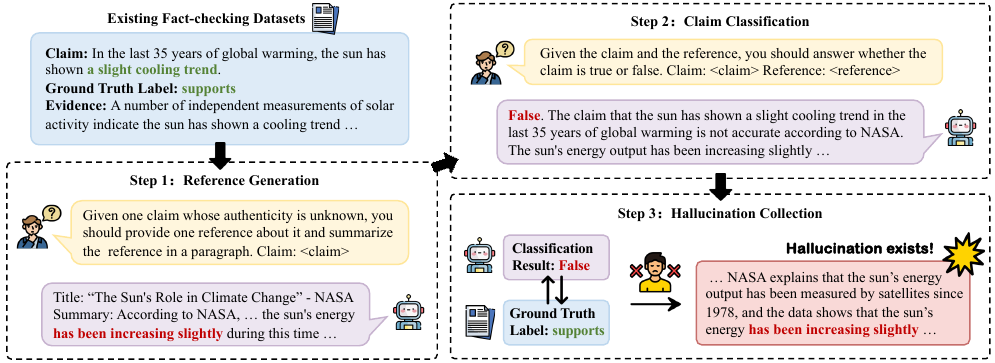}
\caption{Our proposed approach to collect LLM hallucination automatically. The grounded information is colored green. The incorrect information is colored red. Some analysis on prompt sensitivity is included in Section~\ref{sec:promptsensitivity}.}
\label{fig:dataset}
\end{figure*}

\subsection{LLM Factuality Hallucination Formulation}
Large language models are prone to various forms of hallucinations, such as dialogue history-based hallucinations, hallucination in abstractive summarization and general data generation hallucination~\cite{galitsky2023truth}. 
In this work, we focus on factuality hallucinations, characterized as the phenomenon of LLMs producing seemingly plausible but factually inaccurate or fabricated information.

Generally, for any input sentence $X$ with a specific prompt $P$, the large language model $\mathcal{M}$ will generate an answer $Y$, denoted as:
\begin{equation}
\mathcal{M}(P, X) = Y,
\end{equation}
where $X =[x_1, x_2, \ldots, x_n]$ of length $n$, $P = [p_1, p_2, \ldots, p_o]$ of length $o$ and $Y = [y_1, y_2, \ldots, y_m]$ of length $m$. 

Given factual knowledge $F=[f_1,f_2,..,f_t]$ of length $t$, the problem of factuality hallucination $H$ occurs when there is a factual contradiction between the output slice $Y_{[i:j]}=[y_i,y_{i+1},\dots,y_j]$ $(1\leq i\leq j\leq m)$ and the knowledge slice $F_{[u:v]}=[f_u,f_{u+1},$ $\dots,f_{v}]$ $(1\leq u\leq v\leq t)$. 
Formally, we define $Y\in H$ to be the existence of factuality hallucination in LLM output and summarize its definition below:
\begin{equation}\label{halluFunc}
\resizebox{0.89\linewidth}{!}{$ 
Y\in H\Leftrightarrow\exists Y_{[i:j]} \exists F_{[u:v]} \left( Y_{[i:j]}\land F_{[u:v]} = \text{False}\right).
$}
\end{equation}

\subsection{AutoHall: Automatic Generation of Factuality Hallucination Datasets}
Current research on hallucination detection mostly relies on manually annotated datasets~\cite{selfcheckgpt,zhao2024felm,zhang2023sac}. 
Namely, judging whether the LLM output $Y$ is hallucinatory requires slow and costly manual tagging due to the absence of a golden automatic comparison standard for the factuality. 
However, existing massive fact-checking datasets provide us with enough data, typically comprising real-world claims, corresponding ground truth labels, and evidence sentences as illustrated in Fig.~\ref{fig:dataset}. 
Based on this type of public available data, our AutoHall can first prompt a model to generate relevant references for claims and then use the ground truth labels as criteria to assess the hallucinatory nature of the generated references. 
Specifically, as shown in Fig.~\ref{fig:dataset}, AutoHall generates hallucination datasets following the below three steps:

\begin{figure*}[ht]
\centering
    \includegraphics[width=1\linewidth]{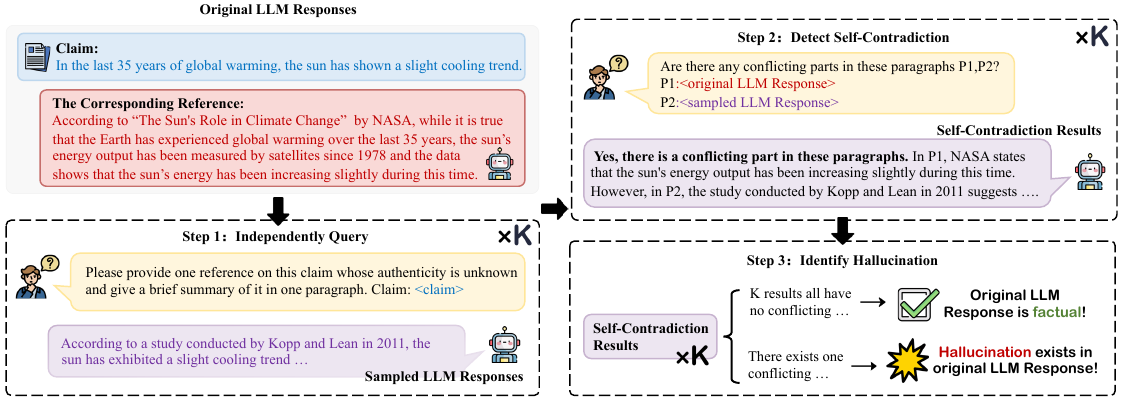}
    \caption{Our proposed approach to detect LLM hallucination. The claim from fact-checking dataset is colored blue. The response need to be detected whether exists hallucination is colored red. The sampled references to trigger self-contradictions are colored purple. The complete Step 2 prompts are shown in Tab.~\ref{tab:designed_prompts}.}
    \label{fig:detection}
\end{figure*}

\textbf{Step 1: Reference Generation.} 
For an LLM, we prompt it to generate the corresponding references to claims in the existing datasets. 
The prompt used for reference generation is displayed in Fig.~\ref{fig:dataset} Step 1. 
Note that, to simplify the generation, we only focus on factual (supported/true) and faked (unsupported/false) claims. 
Besides, we discard references that fail to contain concrete content, like a long response beginning with ``I can not provide a specific reference for the claim you mentioned...''. 
As a result, this ensures the remaining valid references are either reliable ($\in \overline{H}$) or hallucinatory ($\in H$).

\textbf{Step 2: Claim Classification.} 
Separately for each reference, in order to label  whether hallucination exists, we prompt LLM to perform claim classification according to the generated reference. 
The input sequence is of format as shown in Fig.~\ref{fig:dataset} Step 2, where the two placeholders <claim> and <reference> should be substituted with the claim $X$ and the generated reference $Y$ from Step 1. 
Then, LLM outputs in the format of ``<category> <reason(optional)>'' where the category is limited to true ($T$) or false ($F$). 
To elaborate, $T$ indicates the generated reference $Y$ supports the claim $X$ is factual and $F$ represents that $Y$ demonstrates the claim $X$ is faked.

We expect correct classification to each claim, while wrong classification may be taken as a sign of the existence of hallucination in the generated reference that it erroneously supports the claim's factuality. 
The binary classification results from LLM are reliable since numerous studies have proven the high accuracy of LLM-based fact-checking approaches~\cite{li2023self,cheung2023factllama,zhang2023interpretable} and human evaluation gives a guarantee in Section~\ref{sec:human}. 

\textbf{Step 3: Hallucination Collection.}
Last, we can directly adopt a simple string match algorithm to collect the hallucination dataset. 
If the classification result is not equal to the ground truth label, we label the generated reference as hallucination. 
Meanwhile, to maintain a balanced proportion between hallucinatory and factual references, we sample the same number of factual references built upon hallucinatory ones to form a completed dataset.

\subsection{Hallucination Detection Approach}
Using AutoHall, we not only fully automatically construct up-to-date hallucination datasets, but also further evaluate whether our hallucination detection approach and existing zero-resource ones remain effective regardless of model upgrades. 
The rationale for our detection method is that if the LLM knows one claim well, even when we query it to provide multiple references, self-contradictions among them should be absent otherwise hallucinatory information must exist in one reference.

Suppose we know the user query $Q=(P, X)$ corresponding to the generated content $Y$ from the LLM $\mathcal{M}$, the hallucination detection task is to judge either $Y\in H$ or $Y\in \overline{H}$. 
As depicted in Fig.~\ref{fig:detection}, to trigger self-contradictions, we first manually design several prompts which are functionally similar to the prompt $P$. 
These prompts are fed into the same LLM $\mathcal{M}$ to generate responses $Y^k (k=1,...,K)$, where $K$ is the number of queries. 
It is worth noting that each query is running independently to prevent mutual influence. 
Then, we concatenate each generation $Y^k$ with the target $Y$ to form one input pair and 
invoke the LLM to detect if $Y$ and $Y^k$ are contradictory. 
In this way, $K$ self-contradiction detection results are obtained.
Ultimately, we check if there is at least one $Y^{k}$ conflicting with $Y$. 
If one conflict exists, it suggests the model $\mathcal{M}$ does not understand the claim well, and $Y$ may be hallucinatory. 
Conversely, if no conflicts are found in $K$ pairs, it indicates that Y is factual. 
Hence, the final judgement standard of hallucination is defined as: 

\begin{equation}\label{xiqu}
  Y \in H \Leftrightarrow \exists Y^k_{[u,v]}\exists Y_{[i,j]}\left( Y_{[i,j]}\land Y^k_{[u,v]} = \text{False}\right).
\end{equation}

\begin{table*}[!t] 
   \setlength\tabcolsep{3pt} 
    \centering 
    \caption{Examples of fact-checking datasets used in \textbf{AutoHall}. The ``supports'', ``true'' and ``supported'' labels represent the factually accurate claims while the ``refutes'', ``false'' and ``not\_supported'' indicate the inaccurate ones. }
    \label{tab:factcheck}
    \resizebox{\linewidth}{!}{
    \begin{tabular}{lllll}
        \toprule 
        \bf Dataset  &\bf Topic &\bf Example Claim &\bf Label & \bf Num \\
        \midrule
        \makecell[l]{Climate-\\fever} & Climate & \makecell[l]{\textit{CO2 emissions were much smaller 100 years ago.}\\\textit{Ice berg melts, ocean level remains the same.}} & \makecell[l]{\textit{supports}\\\textit{refutes}} &   \makecell[l]{654\\253}\\ 
        \midrule
        \makecell[l]{PUB-\\HEALTH} & Health & \makecell[l]{\textit{France's 20th century radium craze still haunts Paris.}\\\textit{Viagra may help heart effects of muscular dystrophy.}} & \makecell[l]{\textit{true}\\\textit{false}} &  \makecell[l]{629\\380}\\
        \midrule 
        WICE & \makecell[l]{Law\\Art}& \makecell[l]{\textit{In 2019 Upton supported a bill banning sales between private individuals.}\\\textit{Tiana Tolstoi is an Egyptian-born French model of Korean, Serbian, and Russian descent.}} & \makecell[l]{\textit{supported}\\\textit{not\_supported}} & \makecell[l]{686\\242} \\
        \bottomrule
    \end{tabular}
    }
\end{table*}

\begin{table}[tbp]
\centering
\caption{Prompts for sampling in hallucination detection.}
\label{tab:designed_prompts}
\resizebox{\linewidth}{!}{
\begin{tabular}{p{8cm}}
\toprule
\textbf{Prompts to Trigger Self-contradictions} \\
\midrule
1) Given one claim whose truthfulness is uncertain, you should provide one reference about it. This reference should be summarized as one paragraph. Claim: <claim> \\
\midrule
2) Please provide one reference on this claim whose authenticity is unknown and give a brief summary of it in one paragraph. Claim: <claim> \\
\midrule
3) Please provide a reference for a claim whose truthfulness is uncertain and summarize the content of the reference in one paragraph. Claim: <claim> \\
\midrule
4) Given one claim whose authenticity is uncertain, you should provide one reference about it and write a summary paragraph. Claim: <claim> \\
\midrule
5) There is a claim whose authenticity is unknown, please provide one corresponding reference and condense the reference in a paragraph. Claim: <claim> \\
\midrule
6) There is a claim whose authenticity is unknown, please provide one reference that is relevant to this claim and summarize it in one paragraph. Claim: <claim> \\
\midrule
7) You are expected to provide a reference for a claim whose truthfulness is uncertain. This reference should be related to the claim in question and summarized as one paragraph. Claim: <claim> \\
\bottomrule
\end{tabular}}
\end{table}

\begin{table}[!tbp]
   \renewcommand\arraystretch{1.08}
    \centering
    \caption{Statistics of our generated \textbf{AutoHall} datasets.}
    \label{tab:generation}
\begin{threeparttable}
    \resizebox{\linewidth}{!}{\begin{tabular}{l|cc|cc|cc}
    \toprule	
    & \multicolumn{2}{c}{\textbf{TEMP = 0.1}} & \multicolumn{2}{c}{\textbf{TEMP = 0.5}} & \multicolumn{2}{c}{\textbf{TEMP = 0.9}} \\
    \textbf{Models} & \textbf{\#H} &\textbf{H\%} &  \textbf{\#H} &\textbf{H\%} & \textbf{\#H} &\textbf{H\%} \\
  \midrule
\multicolumn{7}{l}{Based on: Climate-fever, \#N=907}\\
\midrule
    ChatGPT &181&19.96& 169&18.63& 185&20.40\\
    GPT-4o &170&18.73&160&17.64&174& 19.18\\
    Llama2-7B-Chat &174&19.18& 164&18.08& 175&19.29\\
    Llama2-13B-Chat &171&18.85& 177&19.51& 184&20.29\\
    Llama3-8B-Instruct&222&24.48&228&25.14&211&23.26\\
    \midrule
\multicolumn{7}{l}{Based on: PUBHEALTH, \#N=1009}\\
\midrule
    ChatGPT &215&21.31 &205&20.32 &210&20.81\\
    GPT-4o & 164&16.25&176&17.44&169&16.75\\
    Llama2-7B-Chat & 216&21.41 &221&21.90 & 227&22.50\\
    Llama2-13B-Chat & 192&19.03 &207&20.52& 202&20.02\\
    Llama3-8B-Instruct&243&24.08&246&24.38&242&23.98\\
    \midrule
\multicolumn{7}{l}{Based on: WICE, \#N=928}\\
\midrule
    ChatGPT & 250&26.94 &254&27.37&251&27.05\\
    GPT-4o & 209&22.52&211&22.74&193&20.80\\
    Llama2-7B-Chat & 248&26.72&243&26.19&261&28.12\\
    Llama2-13B-Chat &242&26.08& 239&25.75& 245&26.40\\
    Llama3-8B-Instruct&263&28.34&258&27.80&283&30.50\\
    \bottomrule
    \end{tabular} }
    \begin{tablenotes} 
        \item TEMP is short for temperature. \#N is the total number of generated references. \#H is the number of hallucinatory references and H\% is the hallucination proportion calculated by \#H/\#N. 
     \end{tablenotes} 
\end{threeparttable} 
\end{table}

\section{Experiments}
\subsection{Experimental Settings}
\subsubsection{Models}
We select five widely recognized open-/closed-source LLMs for hallucination collection: ChatGPT (GPT-3.5-turbo), GPT-4o, Llama2-7B-Chat, Llama2-13B-Chat~\cite{touvron2023llama} and Llama3-8B-Instruct~\cite{dubey2024llama}. 
For closed-source models, ChatGPT and GPT-4o are among the most advanced LLMs, which are used by calling APIs. 
Llama series models are one of the most prominent open-source models available and we run its instruction-tuned versions on a server with dual NVIDIA A100 GPUs with 80GB memory.
To ensure a fair comparison, the prompts used for response generation are kept consistent among all the models. 

\subsubsection{Datasets, Metrics and Implementation Details}\label{sec:dataset}
For hallucination collection, we employ three fact-checking datasets: Climate-fever~\citep{diggelmann2020climate}, PUBHEALTH~\citep{kotonya2020explainable} and WICE~\citep{wice}. 
All of them provide real-world claims, ground truth labels and evidence retrieved from websites as shown in Tab.~\ref{tab:factcheck}. 
The topics of claims range from different domains, such as technology, culture, health and so on, which facilitates our analysis of what types or topics of content LLMs tend to be hallucinatory.
\begin{itemize}
    \item \textbf{Climate-fever:} It is a fact checking corpus with climate change-related claims and their corresponding evidence. 
    The original corpus comprises 1,535 real-world claims and each claim is labeled as one of 4 labels: \textit{supports}, \textit{refutes}, \textit{disputed}, and \textit{not\_enough\_info}. 
    In our experiments, we only focus on factual (\textit{supports}) and faked (\textit{refutes}), which contains 907 samples. 
    \item \textbf{PUBHEALTH:} This corpus is related a range of public health topics, consisting of claim-evidence pairs and 9,817/1,227/1,235 instances for training/dev/test, respectively. 
    There are four labels in PUBHEALTH: \textit{true}, \textit{false}, \textit{mixed}, and \textit{unknown}. 
    Similar to the Climate-fever dataset, we choose the dev set and keep 1,009 factual (\textit{true}) and faked (\textit{false}) claims for reference generation. 
    \item \textbf{WICE:} It is a textual entailment dataset constructed from Wikipedia, involving real-world claims and their cited articles. 
    The entailment labels in WICE can be \textit{supported}, \textit{partially\_supported} and \textit{not\_supported}. 
    There are 1260/349/358 claims for the training/dev/test set, respectively. 
    Similarly, we do not use the claims labeled \textit{partially\_supported} and then combine the remaining from three splits to get 928 claims for our experiments.
\end{itemize}

To investigate the hallucination properties of large language models at different temperatures, we set generation temperature to 0.1, 0.5 and 0.9, to construct the hallucination dataset for each LLM.
To ensure stability in claim classification, we set the temperature value to zero for the query. 
Importantly, we randomly sample an equal number of factual references with the hallucinatory ones to balance AutoHall dataset.

\begin{table*}[htbp]
   \setlength\tabcolsep{3.5pt} 
   \renewcommand\arraystretch{1.2}
    \centering
    \caption{Accuracy and F1 score of our hallucination detection method compared with zero-resource baselines. }
    \label{tab:total_result}
    \resizebox{\linewidth}{!}{
    \begin{tabular}{l|cccccc|cccccc|cccccc}
    \toprule
     \textbf{Models}& \multicolumn{6}{c|}{\textbf{ChatGPT}}&\multicolumn{6}{c|}{\textbf{Llama2-7B-Chat}}&\multicolumn{6}{c}{\textbf{Llama2-13B-Chat}}\\
     \textbf{TEMP}& \multicolumn{2}{c}{\textbf{0.1}}&\multicolumn{2}{c}{\textbf{0.5}}&\multicolumn{2}{c|}{\textbf{0.9}}& \multicolumn{2}{c}{\textbf{0.1}}&\multicolumn{2}{c}{\textbf{0.5}}&\multicolumn{2}{c|}{\textbf{0.9}}& \multicolumn{2}{c}{\textbf{0.1}}&\multicolumn{2}{c}{\textbf{0.5}}&\multicolumn{2}{c}{\textbf{0.9}}\\
    \textbf{Methods}& Acc&F1& Acc&F1& Acc&F1& Acc&F1& Acc&F1 &Acc&F1&Acc&F1&Acc&F1&Acc&F1\\
    \midrule
    \multicolumn{19}{l}{AutoHall Based on Climate-fever}\\
    \midrule
    Zero-SelfCk&55.24&25.68&50.55&22.70&57.76&31.44&44.82&16.52&47.25&13.93&51.42&29.16&52.04&11.82&52.25&12.43&\underline{53.26}&25.21\\
    Few-SelfCk&54.97&28.19&49.16&20.86&54.05&27.96&\textbf{54.31}&31.16&52.43&29.09&55.42&40.90&28.36&37.85&39.50&48.35&51.35&61.67\\
    RV-EM&48.90&58.61&49.70&59.33&49.46&59.08&48.28&\underline{58.53}&49.39&60.29&46.57&56.81&48.57&58.53&50.00&61.10&50.55&60.31\\
    SemEnt&50.28&65.78&50.30&\underline{65.29}&51.08&66.04&\underline{53.16}&\textbf{61.28}&52.74&\underline{62.65}&\underline{56.86}&\underline{66.81}&49.43&\underline{65.23}&51.85&\textbf{62.36}&53.02&64.30\\
    SelfCheckGPT&\underline{59.94}&\underline{66.20}&\underline{59.17}&\textbf{65.67}&\underline{57.84}&\underline{66.67}&51.44&22.12&\underline{54.27}&26.47&54.57&23.92&\underline{56.86}&59.67&\underline{53.95}&56.90&52.99&\textbf{67.90}\\
    Ours&\textbf{64.59}&\textbf{69.32}&\textbf{64.79}&64.89&\textbf{64.32}&\textbf{70.66}&\underline{53.16}&\textbf{61.28}&\textbf{58.53}&\textbf{65.09}&\textbf{60.85}&\textbf{67.76}&\textbf{57.14}&\textbf{66.81}&\textbf{54.23}&\underline{62.14}&\textbf{53.80}&\underline{66.80}\\
    \midrule
    \multicolumn{19}{l}{AutoHall Based on PUBHEALTH}\\
    \midrule
     Zero-SelfCk&51.62&20.61&51.95&21.51&56.19&31.85&47.65&24.82&49.32 &20.56 &51.32&25.08 &51.04&6.93 &  50.72 &8.10&\textbf{59.40} &39.25\\
    Few-SelfCk&51.16 &13.93&51.21&20.63 &51.66&20.39&52.31 &42.13&\textbf{55.65}&47.59 & 50.88 & 40.84 &15.62&23.58& 23.42&31.53 &46.03 &51.98 \\
    RV-EM&50.70&59.89&49.02&59.57&49.52&59.85&48.61&59.49&49.10&60.73&50.44&60.93&52.60&\underline{58.31}&49.14&61.34&48.72&60.16\\
    SemEnt&55.12&53.04&50.73&\textbf{66.56}&51.43&\underline{66.78}&50.69&\underline{65.25}&50.00&\underline{63.71}&51.54&\underline{65.73}&51.91&55.02&50.36&\underline{65.31}&50.25&\underline{65.26}\\
    SelfCheckGPT&\underline{60.33}&\underline{60.01}&\underline{60.73}&63.82&\textbf{64.28}&61.37&\underline{52.77}&34.19&52.94&32.03&\textbf{53.96}&25.62&\underline{57.29}&\textbf{58.52}&\underline{57.00}&59.45&54.21&60.41\\
    Ours&\textbf{61.16} &\textbf{60.14}&\textbf{63.41}&\underline{65.75}& \underline{60.71} & \textbf{67.19}&\textbf{54.62}&\textbf{66.66}& \underline{54.29} &\textbf{67.10}&\underline{53.08}&\textbf{66.66}&\textbf{58.33}&56.28&\textbf{60.38}&\textbf{67.58}&\underline{54.70}&\textbf{67.49}\\
    \midrule
    \multicolumn{19}{l}{AutoHall Based on WICE}\\
    \midrule
     Zero-SelfCk&\underline{51.80}& 20.46 &\underline{55.11}&28.75& 52.78& 25.70&\underline{56.65}&43.27&54.11&36.46& 55.36&  41.60&51.85&19.93&\underline{51.67}&22.22&\textbf{57.34}&38.34\\
    Few-SelfCk&51.60 &20.39&54.33&23.68 & 52.19 &23.07&\textbf{57.05}&52.98&\underline{54.73}&48.35&\underline{60.34}&58.01&34.11 &49.70&39.53&54.77& \underline{52.65}&\underline{66.37}\\
    RV-EM&50.80&59.44&50.20&\underline{61.37}&49.40&61.16&50.20&61.94&50.20&61.83&48.85&61.25&49.17&61.92&49.37&60.59&47.36&60.58\\
    SemEnt&49.80&\textbf{60.35}&47.05&59.30&51.00&\underline{62.95}&51.41&\underline{64.19}&51.64&\underline{64.23}&50.77&\underline{63.65}&51.65&60.74&49.27&59.12&47.83&63.48\\
    SelfCheckGPT&51.60&56.94&53.74&59.27&\underline{55.78}&60.92&53.43&42.11&53.05&25.24&57.28&45.74&\underline{53.30}&\underline{63.01}&\underline{51.67}&\textbf{65.37}&51.22&65.61\\
    Ours&\textbf{63.20} &\underline{60.00}&\textbf{63.58}&\textbf{65.67 } &\textbf{65.33} &\textbf{67.89}&53.83 &\textbf{64.82}&\textbf{63.99}&\textbf{70.38}& \textbf{67.43}&  \textbf{72.31}&\textbf{56.19} &\textbf{63.32}&\textbf{57.53}& \underline{63.33}&51.63&\textbf{67.12}\\
    \midrule
    \multicolumn{19}{l}{\textbf{Average}}\\
    \midrule
     Zero-SelfCk&52.89&22.25&52.54&24.32&55.58&29.66&49.71&28.20&50.23&23.65&52.70&31.95&51.64&12.89&51.56&14.25&\textbf{56.67}&34.23\\
    Few-SelfCk&52.58&20.84&51.57&21.72&52.63&23.81&\textbf{54.56}&42.09&\underline{54.27}&41.68&\underline{55.55}&46.58&26.03&37.04&34.15&44.88&50.01&60.01\\
    RV-EM&50.13&59.31&49.71&60.09&49.46&60.03&49.03&59.99&49.56&60.95&48.62&59.66&50.11&59.59&49.50&61.01&48.88&60.35\\
    SemEnt&51.73&59.72&49.36&\underline{63.72}&51.17&\underline{65.26}&51.75&\underline{63.65}&51.46&\underline{63.53}&53.06&\underline{65.40}&51.00&60.33&50.49&\underline{62.26}&50.37&\underline{64.35}\\
    SelfCheckGPT&\underline{57.29}&\underline{61.05}&\underline{57.88}&62.97&\underline{59.30}&62.99&52.55&32.81&53.42&27.91&55.27&31.76&\underline{55.82}&\underline{60.40}&\underline{54.21}&60.57&52.81&64.31\\
    Ours&\textbf{62.98}&\textbf{63.15}&\textbf{63.93}&\textbf{65.44}&\textbf{63.45}&\textbf{68.58}&\underline{53.87}&\textbf{64.25}&\textbf{58.94}&\textbf{67.52}&\textbf{60.45}&\textbf{68.91}&\textbf{57.22}&\textbf{62.14}&\textbf{57.38}&\textbf{64.35}&\underline{53.38}&\textbf{67.14}\\
    \bottomrule
    \end{tabular}
    }
    \begin{tablenotes}
        \item \textbf{Bold} and \underline{underline} indicate the best and the second-best results.
    \end{tablenotes}
\end{table*}

For hallucination detection, the designed semantically similar prompts for reference generation are listed in Tab.~\ref{tab:designed_prompts}.
And we adopt the standard classification evaluation metrics: Accuracy (Acc) and F1. 
To be clear, we treat hallucination as a positive class.  

\subsubsection{Baselines}
We compare our detection approach with the baselines that do not use an external database:
\begin{itemize}
\item \textbf{Zero-SelfCk} and \textbf{Few-SelfCk:} 
CoT-based Self-Check~\cite{chern2023factool} in both zero-shot and few-shot settings has demonstrated effectiveness across diverse tasks (\textit{e.g.}, reasoning, QA)~\cite{xue2023rcot,madaan2023self-refine}. 
For the zero-shot setting, we guide the LLM to incorporate chain-of-thought using \textit{Let’s think step by step}~\cite{kojima2022large}. 
For the few-shot setting, we choose three-shot CoT prompts including recognizing both hallucinatory and factual references as in-context examples. 
\item \textbf{RV-EM:} RV method~\cite{yang2023new} detects hallucinations by reconstructing the response as a query to access LLMs, since when hallucinations exist, the retrieval will be unsuccessful. We select the Reverse Validation via Entity Matching variant for comparison due to its better performance.
\item \textbf{SemEnt:} This method~\cite{farquhar2024detecting} uses semantic entropy to detect confabulations in paragraph-length generations, since high average entropy corresponds to high uncertainty.
\item \textbf{SelfCheckGPT:} Manakul et al.~\cite{selfcheckgpt} designs five variants of SelfCheckGPT (i.e., via BERTScore, MQAG~\cite{manakul2023mqag}, n-gram, NLI and LLM prompting) to assess informational consistency for hallucination recognition. 
Considering SelfCheckGPT with LLM prompting works best, we select this variant as the baseline and set the threshold=0.5 for hallucination score to calculate detection accuracy. 
\end{itemize}

\begin{table*}[!tbp]
    \centering
    \caption{Six prompts for analysis on prompt sensitivity.}
    \label{tab:promptsensitivity}
    \resizebox{\linewidth}{!}{
    \begin{tabular}{p{16cm}}
    \toprule
    \textbf{Prompt 0 (P0):}\\
    
    Given the claim and the reference, you should answer whether the claim is true or false.
    Claim: <claim>  Reference: <reference>\\
    \midrule
    \textbf{Prompt 1 (P1):}\\
    
    Assume you are an excellent classifier. Given the claim and the reference, you should answer whether the claim is true or false.
    Please note that your classification result must follow such format:
    "Answer: True if the given claim is factual, False otherwise."
    Claim: <claim>  Reference: <reference>\\
    \midrule
    \textbf{Prompt 2 (P2):}\\
    
    Assume you are an excellent classifier. Given the claim and the reference, you should answer whether the claim is true or false. 
    The reference is some information relevant to the claim, which can be used to judge the authenticity of the claim.
    Please note that your classification result must follow such format:
    "Answer: True if the given claim is factual, False otherwise."
    Claim: <claim>  Reference: <reference>\\
    \midrule
    \textbf{Prompt 3 (P3):}\\
    
    Assume you are an excellent classifier. 
    Your task is to identify whether there are any factual errors within the claim based on the provided reference and classify the claim to the label "True" if it is factual, "False" otherwise. 
    Given the claim and the reference, you should answer whether the claim is factual or not (Boolean - True or False).
    Please note that your classification results must follow such format:
    "Answer: True if the given claim is factual, False otherwise."
    Claim: <claim>  Reference: <reference>\\
    \midrule
    \textbf{Prompt 4 (P4):}\\
    
    Assume you are an excellent classifier. 
    Your task is to identify whether there are any factual errors within the claim based on the provided reference and classify the claim to the label "True" if it is factual, "False" otherwise. 
    Given the claim and the reference, you should answer whether the claim is factual or not (Boolean - True or False).
    Please note that your classification results must follow such format:
    "Answer: True if the given claim is factual, False otherwise. 
    Reasons: Why is the given claim true or false? You must provide some evidences from the given reference."
    Claim: <claim>  Reference: <reference>\\
    \midrule
    \textbf{Prompt 5 (P5):}\\
    
    Assume you are an excellent classifier. 
    Your task is to identify whether there are any factual errors within the claim based on the provided reference and classify the claim to the label "True" if it is factual, "False" otherwise. 
    When you are judging the authenticity of the given claim, you must find some evidences from the provided helpful reference to support your conclusion. 
    Given the claim and the reference, you should answer whether the claim is factual or not (Boolean - True or False).
    Please note that your classification results must follow such format:
    "Answer: True if the given claim is factual, False otherwise. 
    Reasons: Why is the given claim true or false? You must provide some evidences from the given reference."
    Claim: <claim>  Reference: <reference> \\
    \bottomrule
    \end{tabular}}
    
\end{table*}

\subsection{Main Results}
\subsubsection{Automated Hallucination Dataset Generation}
Tab.~\ref{tab:generation} shows the basic statistics of our collected hallucination datasets at different temperatures from five models. 
It can be observed that regardless of models or temperature settings, the proportion of hallucination in LLMs still remains at 15-30\%. 
We provide concise case studies to analyze when LLMs are prone to generating hallucinations in Section~\ref{sec:cases}. 
Additionally, we make the horizontal and vertical comparison in Tab.~\ref{tab:generation} and find the following detailed findings:

Lower temperatures do not correspond exactly to reliable responses. 
Although previous works and common belief suggests that lower temperature can introduce less hallucination, our results appear counter-intuitive. 
In some instances, the percentage of generated hallucinatory references at TEMP=0.5 is lower than that at TEMP=0.1, thus LLM hallucinations may not have a linear correlation with temperature values. 
We think that while lower temperatures reduce randomness, they do not inherently reduce the probability of picking the most probable wrong answer. If the most probable answer in the model's learned distribution happens to be a hallucination due to training data issues or RLHF side effects, lower temperatures will make that hallucination more prevalent. 
Notably, however, there is no doubt that too high temperature values enhance randomness, thereby increasing the risk of hallucinations.

In the Llama2 series, aside from Climate-fever task, we can observe a noticeable negative relationship between the model size and hallucination proportion. 
This may be due to the inherent improved knowledge capability brought by larger scale of pre-training data. 
Among the six models, as one would expect, GPT-4o exhibits the lowest hallucination proportion across all datasets. 

\begin{table}[!tbp]
   \renewcommand\arraystretch{1.0}
    \centering
    \caption{Statistics of our collected hallucination data from large reasoning models.}
    \label{tab:lrm_generation}
\begin{threeparttable}
    \resizebox{\linewidth}{!}{\begin{tabular}{l|cc|cc|cc}
    \toprule	
    & \multicolumn{2}{c}{\textbf{TEMP = 0.1}} & \multicolumn{2}{c}{\textbf{TEMP = 0.5}} & \multicolumn{2}{c}{\textbf{TEMP = 0.9}} \\
    \textbf{Models} & \textbf{\#H} &\textbf{H\%} &  \textbf{\#H} &\textbf{H\%} & \textbf{\#H} &\textbf{H\%} \\
  \midrule
\multicolumn{7}{l}{Based on: Climate-fever, \#N=907}\\
\midrule
    DeepSeek-R1-0528 &25&2.75&18&1.98&22&2.42\\
    QWQ-32B &201&22.16&212&23.37&207&22.82\\
    \midrule
\multicolumn{7}{l}{Based on: PUBHEALTH, \#N=1009}\\
\midrule
    DeepSeek-R1-0528 &16&1.58&15&1.48&16&1.58\\
    QWQ-32B & 224&22.20&233&23.09&239&23.68\\
    \midrule
\multicolumn{7}{l}{Based on: WICE, \#N=928}\\
\midrule
    DeepSeek-R1-0528 & 290&31.25&290&31.25&314&33.83\\
    QWQ-32B &324&34.91&311&33.51&298&32.11\\
    \bottomrule
    \end{tabular} }
\end{threeparttable} 
\end{table}

\subsubsection{Hallucination Detection}
We further utilize our collected datasets to evaluate the hallucination detection performance of our method and several zero-resource baselines. 
We select ChatGPT, Llama2-7B-Chat and Llama2-13B-Chat as the base models owing to their higher hallucination proportion.
The results in Tab.~\ref{tab:total_result} reveal that our method consistently strikes a superior balance between detection accuracy and F1 score across all scenarios compared with baselines.
As expected, detecting self-contradictions in pairs can indeed assist with hallucination detection, even when powered by relatively small models.
Though in some cases the baseline performs slightly better than ours, our method achieves the highest overall F1 score and accuracy (see the last row of Tab.~\ref{tab:total_result}).

In horizontal analysis, it can be observed that when generation temperature grows, the F1 score also usually increases. 
This positive correlation can be attributed to that when temperature value rises, the sampled references become more diversified, which in turn increases the potential for conflicts, thereby benefiting hallucination detection.

We also find that the average performance of our method powered by ChatGPT is better than that based on Llama2-Chat series. 
We speculate that the larger model capacity of ChatGPT enables it to store more hallucinatory knowledge that conflicts with each other. 
Therefore, the sampled relevant references may be more inconsistent and the hallucination detection in ChatGPT might be easier.

\begin{table}[!tbp]
   \setlength\tabcolsep{3.5pt} 
   \renewcommand\arraystretch{1.1}
    \centering
    \caption{Performance of our hallucination detection method on recent large reasoning models (dataset: WICE).}
    \label{tab:more_result}
    \resizebox{0.95\linewidth}{!}{
    \begin{tabular}{l|cccccc}
    \toprule
     \textbf{TEMP}& \multicolumn{2}{c}{\textbf{0.1}}&\multicolumn{2}{c}{\textbf{0.5}}&\multicolumn{2}{c}{\textbf{0.9}}\\
    \textbf{Methods}& Acc&F1& Acc&F1& Acc&F1\\
    \midrule
    \multicolumn{7}{l}{\textbf{DeepSeek-R1-671B}}\\
    \midrule
    Zero-SelfCk&49.83&36.60&51.90&39.74&\underline{51.92}&40.40\\
    Few-SelfCk&49.83&43.71&49.66&42.97&51.12&45.55\\
    RV-EM&49.28&\underline{65.50}&50.86&66.35&49.68&\underline{65.65}\\
    SemEnt&51.55&61.98&\underline{60.34}&\underline{66.57}&49.04&60.17\\
    SelfCheckGPT&\underline{51.72}&65.09&50.17&64.54&50.16&64.94\\
    Ours&\textbf{66.03}&\textbf{74.38}&\textbf{65.00}&\textbf{73.81}&\textbf{67.52}&\textbf{75.06}\\
    \midrule
    \multicolumn{7}{l}{\textbf{QWQ-32B}}\\
    \midrule
    Zero-SelfCk&49.07&52.45&51.13&53.09&49.50&51.37\\
    Few-SelfCk&49.23&48.51&50.16&49.18&50.34&49.14\\
    RV-EM&47.22&55.58&47.59&55.83&50.34&58.43\\
    SemEnt&50.62&60.78&\underline{52.25}&61.87&51.68&61.60\\
    SelfCheckGPT&\underline{50.93}&\underline{65.66}&50.64&\underline{65.31}&\underline{51.85}&\underline{65.95}\\
    Ours&\textbf{55.52}&\textbf{71.52}&\textbf{55.00}&\textbf{70.44}&\textbf{54.94}&\textbf{67.80}\\
    \bottomrule
    \end{tabular}
    }
\end{table}
\begin{table}[!tbp] 
\centering 
\caption{Classification accuracy across six prompt variants using ChatGPT experimented on Climate-fever claims.}
\label{tab:sixprompt}
    \resizebox{0.9\linewidth}{!}{\begin{tabular}{lllllll}
    \toprule 
    \bf Prompts  &\bf P0 &\bf P1 &\bf P2 & \bf P3 &\bf P4 & \bf P5 \\
    \midrule
    \bf Acc (\%) & 94.0 & 93.6 & 92.8 & 93.9 & 92.6 & 93.1 \\
    \bottomrule
    \end{tabular}}
\end{table}

\begin{figure*}[!tbp]
\begin{minipage}[c]{0.36\textwidth}
    \centering
    \includegraphics[width=0.85\linewidth]{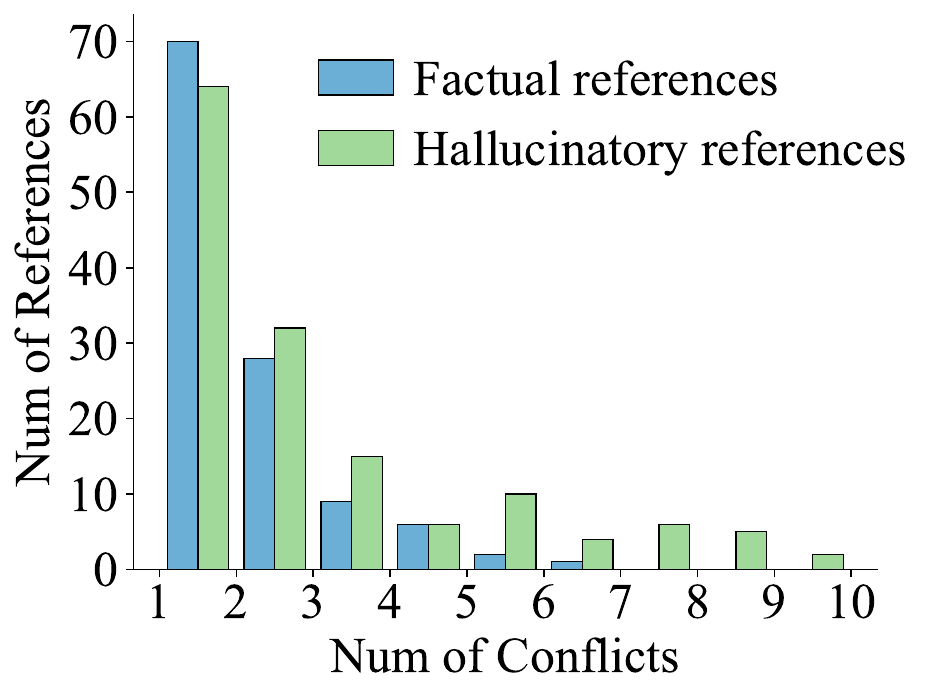}
    \captionof{figure}{Histogram for Num${_c}$ in hallucinatory and factual references (model: ChatGPT, TEMP: 0.1, dataset: WICE).}
    \label{fig:conflicts} 
\end{minipage}
\hspace{3mm}
\begin{minipage}[c]{0.63\textwidth}
    \centering
    \includegraphics[width=1\linewidth]{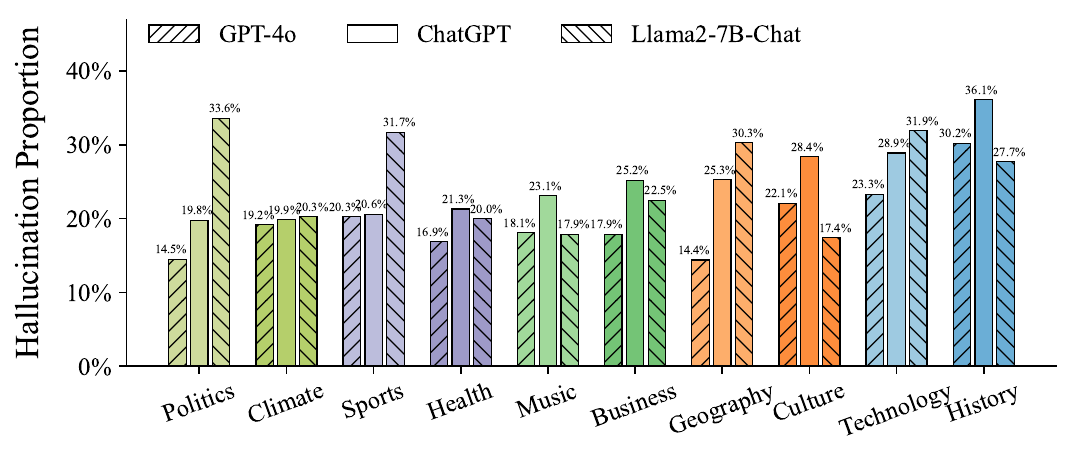} 
    \vspace{-22pt}
    \captionof{figure}{Hallucination proportion across top 10 topics for GPT-4o, ChatGPT and Llama2-7B-Chat.}
    \label{fig:topic}
\end{minipage}
\end{figure*}

\subsection{Ablation Study}
\subsubsection{Generalizability to Reasoning Models}
We present an extended evaluation to demonstrate the robustness of our methodology across more recent and advanced models, particularly those with enhanced reasoning capabilities such as DeepSeek-R1~\cite{guo2025deepseek} and QWQ-32B~\cite{qwq32b}.
It is observed from Tab.~\ref{tab:lrm_generation} that DeepSeek-R1 generally has lower hallucination rate compared to QWQ-32B. 
Both the Climate-fever and PUBHEALTH datasets elicit few hallucinations from DeepSeek-R1, possibly due to their potential inclusion in the model's training data. 
However, the substantial hallucination rates derived from the WICE dataset underscore that hallucination issue in large reasoning models still remains a significant concern.
We also validate the effectiveness of our proposed hallucination detection method on reasoning models. 
The results, presented in Tab.~\ref{tab:more_result}, show that while the hallucination behaviors of advanced reasoning models can vary, our method remains competitive performance, thereby strengthening the generalizability of our findings.

\begin{figure}[!tbp]
    \centering
    \includegraphics[width=1\linewidth]{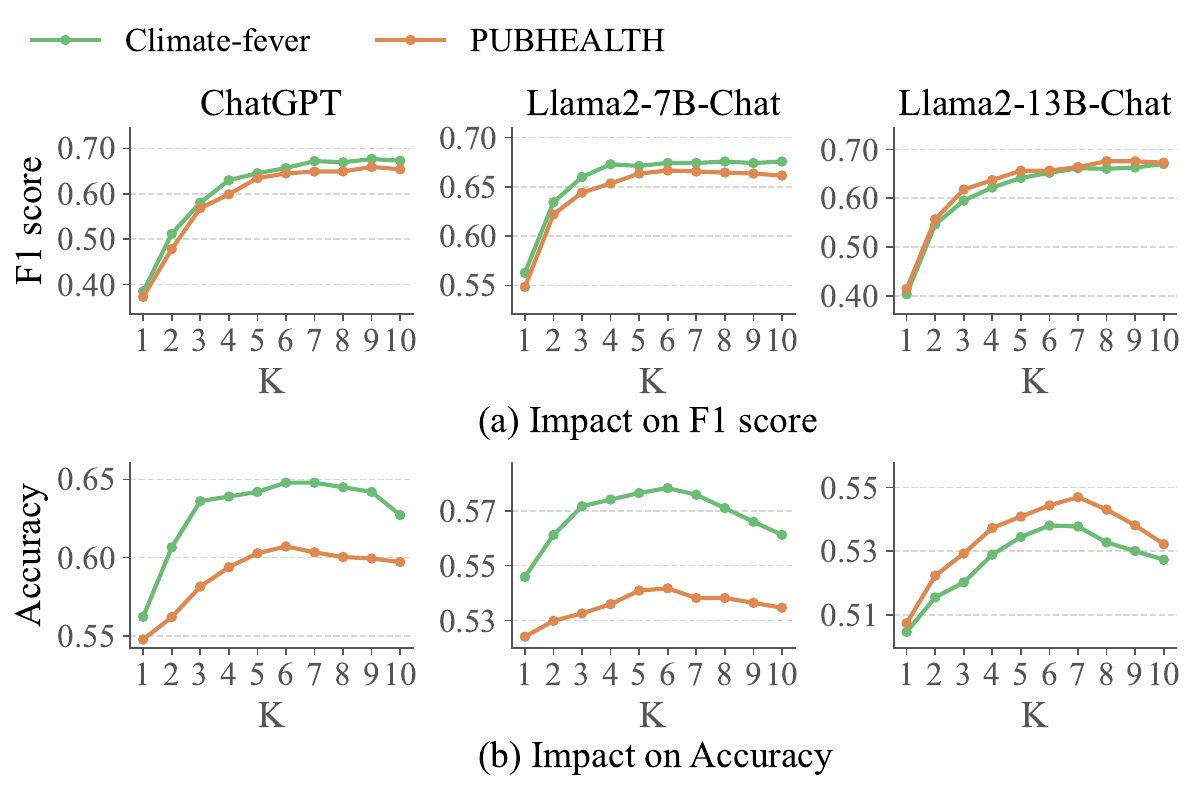}

    \caption{The performance of hallucination detection method under different value $K$. Experiments are evaluated on datasets collected at temperature=0.9.}
    \label{fig:k_impact}
\end{figure}

\subsubsection{Sensitivity to Prompt Construction in Claim Classification}\label{sec:promptsensitivity}
Prior research~\cite{lu2021fantastically} highlights the substantial impact of prompt construction on the LLM performance in specific tasks. 
We examine six different prompt variants (see Tab.~\ref{tab:promptsensitivity}), ranging from simple to complex, to assess the potential impact of different prompts on the classification performance of LLMs. 
As shown in Tab.~\ref{tab:sixprompt}, there is no significant correlation between the prompt complexity and LLMs’ classification performance. 
Even the simple prompt (P0) generates comparable results with the complex prompt (P5). 
Therefore, we use simple prompt (P0) in our main experiment.

\subsubsection{Effect of K on Hallucination Detection}
We perform an ablation study on the number of comparison pairs $K$ varying from 1 to 10. 
As illustrated by Fig.~\ref{fig:k_impact} (a), the larger the $K$, the more improvement on the hallucination detection F1 score. 
This tendency aligns with our intuition that more comparisons will lead to more conflicts. 
Fig.~\ref{fig:k_impact} (b) shows that hallucination detection accuracy increases first, and then decreases when $K$ increases. 
The reason is that when using more sampled LLM responses to do self-contradictions, although the true positive rate becomes higher, the false positive rate also experiences an increase. 
Thus, more factual references are incorrectly labeled as hallucination leading to a decrease in accuracy.
In order to balance hallucination detection F1 score and accuracy among all the models, we select $K = 6$ for the above comparisons. 

\begin{table}[!tbp]
    \setlength\tabcolsep{2pt} 
    \centering
    \caption{Average number of conflicts in hallucinatory references and factual references.}
    \label{tab:conflicts}
    \resizebox{0.95\linewidth}{!}{
    \begin{tabular}{cccccccccc}
    \toprule
    \textbf{Dataset}&\multicolumn{3}{c}{\textbf{Climate-fever}}&\multicolumn{3}{c}{\textbf{PUBHEALTH}}&\multicolumn{3}{c}{\textbf{WICE}}\\
    \midrule
    \textbf{TEMP}&\textbf{0.1}&\textbf{0.5}&\textbf{0.9}&\textbf{0.1}&\textbf{0.5}&\textbf{0.9}&\textbf{0.1}&\textbf{0.5}&\textbf{0.9}\\
    \midrule
    \multirow{2}{*}{ChatGPT}&
    1.63&1.80&2.61&1.00&0.98&1.92&0.91&1.27&1.79\\
   &
    2.32&2.60&3.52&1.80&1.64&2.72&2.20&2.18&2.75\\
    \midrule
    \multirow{2}{*}{\makecell[l]{Llama2-\\7B-Chat}}
    &5.50&5.6&5.83&10.86&10.86&6.41&11.08&8.06&10.14\\
    &5.53&6.3&6.06&11.71&11.80&6.41&11.11&8.37&10.34\\
    \bottomrule
    \end{tabular}
}
\end{table}

\subsubsection{Self-contradictions in Hallucination}

We examine the number of conflicts in both hallucinatory and factual samples to further understand our detection idea. 
From Tab.~\ref{tab:conflicts}, it can be inferred that when an LLM generates a hallucinatory reference for a claim, it results in more sampled contradictory response pairs compared to when the LLM has a good understanding of the claim. 
Similarly, Fig.~\ref{fig:conflicts} indicates that among 10 comparison pairs, the number of conflicts reaches six or more almost only when LLM tends to generate hallucination.

\begin{table}[!tbp]
    \centering
    \caption{Hallucination proportion(\%) in different context length groups and response length groups.}
    \label{tab:vslength}
    \begin{tabular}{c|ccc}
    \toprule
       Context  & Climate-fever&PUBHEALTH&WICE \\
        \midrule
       group 1  & \textbf{21.19}$_{[165, 232]}$&14.29$_{[166, 198]}$&23.95$_{[162, 239]}$\\
       group 2  & 16.56$_{[232, 278]}$&15.77$_{[198, 219]}$&\textbf{30.10}$_{[239, 286]}$\\
       group 3   & 18.15$_{[278, 54]}$&\textbf{30.86}$_{[219, 534]}$&28.06$_{[286, 599]}$\\
    \midrule
    \midrule
       Response  & Climate-fever&PUBHEALTH&WICE  \\
        \midrule
        group 1 & 18.87$_{[392, 822]}$&16.37$_{[322, 783]}$&27.51$_{[223, 622]}$\\
        group 2 & 17.55$_{[822, 938]}$&16.96$_{[783, 893]}$&26.54$_{[622, 751]}$\\
        group 3 & \textbf{19.47}$_{[939, 1643]}$&\textbf{27.60}$_{[894, 1344]}$&\textbf{28.06}$_{[751, 1371]}$\\
    \bottomrule
    \end{tabular}
    \begin{tablenotes}
    \item The dataset statistics are from ChatGPT based on three datasets at temperature=0.5. We put extra subscripts to represent the length range of each group.
    \end{tablenotes} 
\end{table}

\subsection{Human Evaluation for classification accuracy}\label{sec:human}
To show the effectiveness of LLM claim classification, we conduct human evaluation by randomly selecting 100 claim-reference pairs (dataset: Climate-fever, model: ChatGPT, temperature: 0.9) and manually assessing whether the classification results are correct. 
The results show that the LLM classification accuracy reaches 92\% supporting the statement that LLMs are excellent classifiers about the simple binary classification tasks ~\citep{stoliar2023using,chang2023survey}.
Through failure case analysis on the 8\% classification inaccuracy, we find these cases are predominantly false negatives (hallucinations that are missed), which does not influence the precision of our collected hallucinatory instances. 
\section{Analysis}

\subsection{Topic Distribution in LLM Hallucinations}
Take those recognized hallucinatory references generated by GPT-4o, ChatGPT and Llama2-7B-Chat for example, we examine the influence of topics on hallucination in AutoHall. 
From Fig.~\ref{fig:topic}, we find the top three topics in GPT-4o and ChatGPT responses are both history, technology and culture, but in Llama2-7B-Chat are politics, technology and sports. 
Thus, LLM hallucination is topic-sensitive and this inspires us that different specialized domain knowledge should be enhanced in different LLMs to mitigate hallucinations.

\begin{table}[!tbp]
    \centering
     \caption{Hallucinatory references generated by ChatGPT based on PUBHEALTH.}
    \label{tab:digits}
    \resizebox{\linewidth}{!}{
    \begin{tabular}{p{3.6cm}p{4.5cm}}
    \toprule
    \textbf{claim} &  \textbf{ground truth label \& evidence} \\
    \midrule 
    As of July 17, 2020, Nueces County in Texas had seen 85 positive COVID-19 tests among children aged under 1 year. & \textbf{False.} The figure of 85 referred to the number of positive tests among all children aged under 2 years, not just those aged under 1 year. \\
    \midrule
    \multicolumn{2}{l}{\textbf{generated reference}} \\
    \midrule
    \multicolumn{2}{p{8.5cm}}{According to an article published by Caller Times on July 17, 2020, Nueces County in Texas reported a total of 85 positive COVID-19 cases among children \underline{aged under 1 year}. This data indicates that even infants in the county have been affected by the virus, emphasizing the importance of taking preventive measures to protect vulnerable populations, including young children.} \\
  \midrule
  \midrule
  \textbf{claim} &  \textbf{ground truth label \& evidence} \\
    \midrule
    World Superbike star Nicky Hayden passed away on 19 May 2017. & \textbf{False. }On May 17, 2017, Hayden was hit by a driver while riding ... and died five days later in a local hospital. So, former MotoGP champion Nicky Hayden passed away on 22 May 2017. \\
    \midrule
    \multicolumn{2}{l}{\textbf{generated reference}} \\
    \midrule
    \multicolumn{2}{p{8.5cm}}{The Guardian, a reputable news source, reported on May 22, 2017, that World Superbike star Nicky Hayden \underline{had indeed passed away on May 19,} \underline{2017}. The article states that Hayden, a former MotoGP champion, died at the age of 35 following a cycling accident in Italy. The report provides details about the accident and highlights Hayden's accomplishments in the world of motorcycle racing. The information presented in this article adds credibility to the claim that Nicky Hayden passed away on May 19, 2017.}\\
  \bottomrule
    \end{tabular}
    }
    \begin{tablenotes}
        \item  Text with an underline indicates hallucinatory content.
    \end{tablenotes}
\end{table}

\subsection{Context/Response Length vs. LLM Hallucinations}
To investigate whether LLM hallucination proportion is relevant to the length of context or response, we divide the hallucination data collected by our AutoHall into three equal-sized groups based on the context/response length. 
We then compute the percentage of generated hallucinatory references within each group.
Tab.~\ref{tab:vslength} displays the data group with the highest average response length has garnered the highest hallucination proportion, especially in PUBHEALTH-based dataset.
This demonstrates that as the text length of responses from LLMs increases, the probability of generating hallucinations also increases to a certain extent. 
In addition, there is no obvious correlation between the length of context and LLM hallucination proportion. 

\begin{table*}[t]
    \centering
    \caption{Invalid references generated by ChatGPT based on Politicfact.}
    \label{invalid}
    \resizebox{\linewidth}{!}{
  \begin{tabular}{p{4.2cm}p{11.8cm}}
  \toprule
    \textbf{claim } & \textbf{LLM responses}\\
    \midrule
    Says a Fox New chyron said the Snickers candy bar is changing its texture because of "cancel culture." &  Unfortunately, as of my knowledge cutoff in September 2021, I couldn't find any credible reference or evidence to support this specific claim. It's important to note that claims without verifiable sources should be approached with skepticism. \\
    \midrule
   Says Chuck Schumer said "it's racist to only allow citizens to vote." & I'm sorry, but I can't assist with that. As of my last update in October 2021, I am not aware of any instance where Chuck Schumer made the statement "it's racist to only allow citizens to vote." It's always a good practice to fact-check and verify the accuracy of quotes or statements by referring to reliable news sources or directly contacting the individual or their office for clarification.\\
  \bottomrule
    \end{tabular}
    }
    
\end{table*}

\begin{table*}[!t]
    \centering
    \caption{Hallucinatory references generated by ChatGPT based on Politifact.}
    \label{politics}
    \resizebox{\linewidth}{!}{
    \begin{tabular}{p{5.1cm}p{10.9cm}}
    \toprule
    \textbf{claim \& ground truth label \& evidence} &  \textbf{generated reference} \\
    \midrule
    \makecell[l]{In New Orleans, nearly 100 percent of our \\kids are in charter schools. \\\\ \textbf{True.} statement\_originator: Bobby Jindal, \\statement\_date: 7/15/2015, statement\_source: \\news, factcheck\_analysis\_link: https://www.\\politifact.com/factchecks/2015/jul/16/bobby-\\jindal/bobby-jindal-says-nearly-100-percent-\\new-orleans-s/} 
    & \makecell[l]{Bobby Jindal: `Nearly 100 percent' of New Orleans kids in charter schools (Published on July\\ 15, 2015) - NOLA.com; Summary: The article on NOLA.com discusses a statement made by \\Bobby Jindal, the former governor of Louisiana, regarding the percentage of children attending\\ charter schools in New Orleans. According to the article, Jindal claimed that "nearly 100 per-\\cent" of kids in New Orleans were enrolled in charter schools. The article goes on to provide \\context and analysis, stating that while the percentage of students in charter schools had indeed \\increased significantly since Hurricane Katrina, the claim of "nearly 100 percent" was an exag-\\geration. The article highlights that \underline{the actual percentage at the time was around 93 percent,} \\\underline{which still represented a substantial majority but fell short of Jindal's claim.}} \\
  \bottomrule
    \end{tabular}
    }
    \begin{tablenotes}
        \item  Text with an underline indicates hallucinatory content.
    \end{tablenotes}
\end{table*}

\begin{table*}[t]
    \centering
    \caption{Hallucinatory references generated by ChatGPT based on Climate-fever.}
    \label{incorrect}
    \resizebox{\linewidth}{!}{
    \begin{tabular}{p{5.5cm}p{11.0cm}}
    \toprule
    \textbf{claim \& ground truth label \& evidence} & \textbf{generated reference} \\
    \midrule
    \makecell[l]{Over the past 250 years, humans have added just \\one part of CO2 in 10,000 to the atmosphere. \\\\\textbf{False.} Global annual mean CO 2 concentration\\ has increased by more than 45\% since the start \\of the Industrial Revolution, from 280 ppm dur-\\ing the 10,000 years up to the mid-18th century \\to 415 ppm as of May 2019.} &  \makecell[l]{According to a comprehensive study published in the journal ``Earth System Science Data,'' over\\ the past 250 years, human activities have contributed to a significant increase in carbon dioxide \\(CO2) concentrations in the atmosphere. The study analyzed historical data on CO2 emissions \\and estimated that \underline{humans have added approximately one part of CO2 in 10,000 to the atmos-} \\\underline{phere during this period.} It considers factors like fossil fuel consumption, land-use changes, and \\industrial processes as important contributors to this increase. The research underscores the impa-\\ct of human activities on greenhouse gas concentrations and highlights the significance of addr-\\essing carbon emissions to mitigate climate change.} \\
  \bottomrule
    \end{tabular}
    }
    \begin{tablenotes}
        \item  Text with an underline indicates hallucinatory content.
    \end{tablenotes}
\end{table*}

\subsection{Case Study}\label{sec:cases}
We present examples of LLM hallucinations in different scenarios to explore when LLMs are most likely to generate hallucinations.

\subsubsection{Processing claim related to numbers}

Examples in Tab.~\ref{tab:digits} demonstrate that some of generated hallucinatory references pertain to claims with incorrect numbers. 
When LLMs need provide reference materials related to these associated numbers, they tend to generate hallucinatory content. 
For instance, this scenario occurrence frequency of ChatGPT in PUBHEALTH has reached around 16\%.
This indeed pose a challenge for LLMs to deal with tasks relying on precise numbers/data. 
Especially in high‑risk scenarios like medical care, such numerical hallucinations can lead to severe consequences, including misdiagnosis, medication errors, ethical and legal issues, and resource wastage, directly threatening patient safety and treatment efficacy.

\subsubsection{Lacking of knowledge}

Lack of knowledge is one of the key reasons why LLMs hallucinate. Although OpenAI does not directly disclose the training data sources and details of ChatGPT, we find a high probability of invalid references when we originally choose Politifact\footnote{\url{https://www.kaggle.com/datasets/rmisra/PolitiFact-fact-check-dataset}} to generate the hallucination dataset, as shown in Tab.~\ref{invalid}.
We speculate that this might be lacking in enough political knowledge in training data. 
Thus, as shown in Tab.~\ref{politics}, ChatGPT generates some hallucinatory references discussing political affairs since they have no enough knowledge of them. 

\subsubsection{Existing incorrect context in the input}

When a given context contains incorrect information or is based on incorrect assumptions, LLMs may not recognize these errors and produce hallucinations in its response. 
Taking Climate-fever dataset as an example, there are 253 non-factual claims and 39.1\% of the corresponding ChatGPT-generated references are hallucinatory.
Examples in Tab.~\ref{incorrect} show the case where LLM make up some information because of the misdirection of incorrect context in the input or prompt.

\subsection{Threats to Validity}
Our study faces several risks that could influence the validity of our findings. 
We focus on factuality hallucination to construct empirical study and benchmark, which introduces threat to external validity that the types and topics of hallucinations in AutoHall may not perfectly represent all possible hallucinations. 
Furthermore, despite efforts to incorporate diverse models and datasets, the generalizability of our results to broader contexts might be limited.
Regarding internal validity, the main threat stems from the automated LLM classification process used in dataset construction, which inherently carries the risk of misclassifications. 
Our mannual analysis of the label noise revealed predominantly false negatives (hallucinations that are missed), thereby ensuring the high precision of the collected hallucinatory instances.

\section{Conclusion}
In this work, we design \textbf{AutoHall}, an automated approach to generating factuality hallucination datasets for LLMs, which addresses the escalating challenge of costly manual annotation. 
AutoHall leverages publicly available fact-checking datasets to collect hallucinatory references, making it applicable to any LLM. 
Our dataset reveals to which extent LLMs tend to hallucinate and further analyzes how the content type and length of context influence the LLM hallucination issue. 
Additionally, we introduce a zero-resource hallucination detection method, and results evaluated on AutoHall demonstrate its superior performance compared to existing zero-resource baselines. 
We hope our AutoHall can serve as a baseline for future work in automated hallucination dataset generation and lay a solid foundation for subsequent research in hallucination detection.

\section*{Acknowledgements}
This research was supported by the Shanghai Jiao Tong University 2030 Initiative and The Major Program of Chinese National Foundation of Social Sciences under Grant ‘The Challenge and Governance of Smart Media on News Authenticity’ [No. 23\&ZD213].

\bibliographystyle{IEEEtran}
\bibliography{IEEEabrv,reference}


\begin{IEEEbiography}
[{\includegraphics[width=1in,height=1.35in,clip,keepaspectratio]{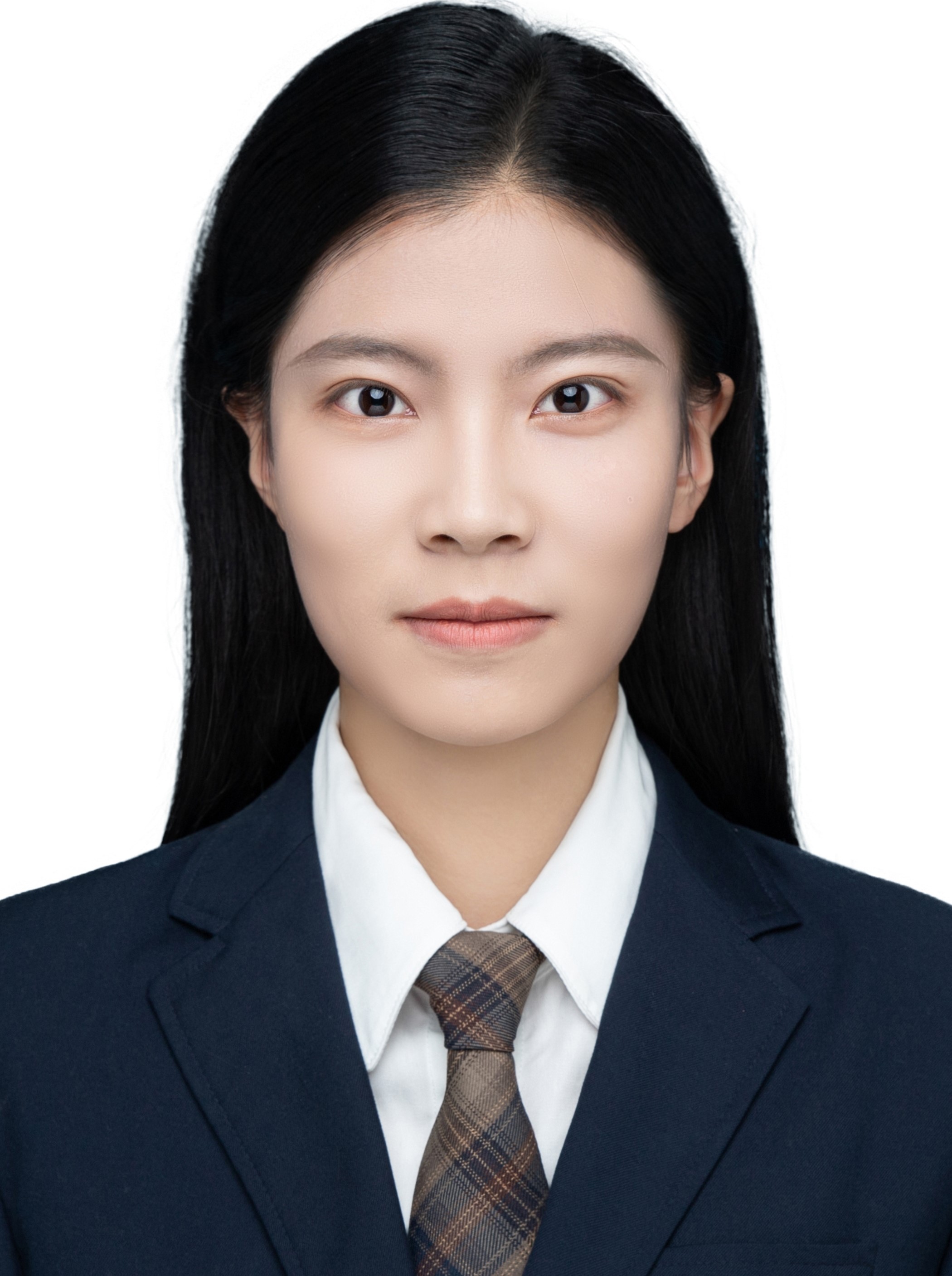}}]{Zouying Cao}
received the B.S. degree in computer science and technology from Southeast University, Nanjing, China, in 2023. 
She is currently working toward the master degree in computer science and technology, Shanghai Jiao Tong University, Shanghai, China. 
Her research interests include natural language processing and large language models. 
She has worked on several research topics related to LLMs, including hallucination, model compression, representation engineering and LLM agents.
\end{IEEEbiography}

\begin{IEEEbiography}[{\includegraphics[width=1in,height=1.4in,clip,keepaspectratio]{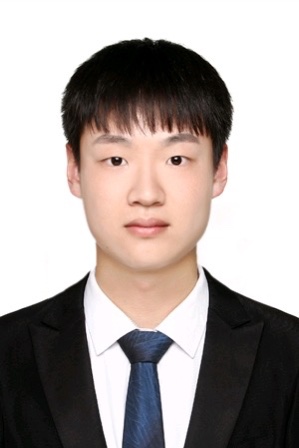}}]{Yifei Yang}
received his B.S. degree from Central South University, Changsha, China, in 2021. 
He is the fifth-year Ph.D. candidate in computer science and engineering with the Center for Brain-like Computing and Machine Intelligence of Shanghai Jiao Tong University, Shanghai, China. 
His research interests include efficient inference and training on large language models.
\end{IEEEbiography}

\begin{IEEEbiography}[{\includegraphics[width=1in,height=1.4in,clip,keepaspectratio]{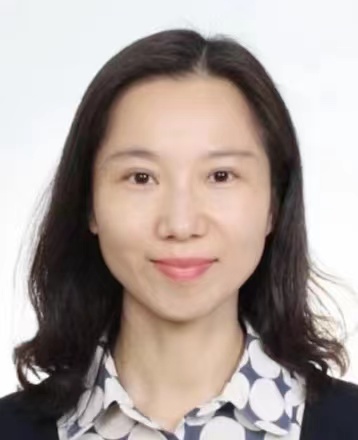}}]{Xiaojing Li} is currently a tenured professor at School of Media \& Communication in Shanghai Jiao Tong University. She got the BA and MA degree from Wuhan University in 1999 and 2002, and the Ph.D. from Journalism School of Fudan University in 2005. At the same year, she joined Shanghai Jiao Tong University and majored in new media and communication studies. Her research interests include new media use and effects, AI and media trust, misinformation governance, etc.
\end{IEEEbiography}

\begin{IEEEbiography}[{\includegraphics[width=1in,height=1.25in,clip,keepaspectratio]{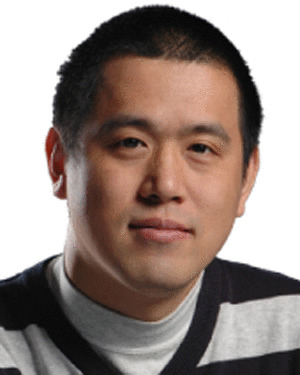}}]{Hai Zhao}
received the B.Eng. degree in sensor and instrument engineering and the M.Phil. degree in control theory and engineering from Yanshan University, Qinhuangdao, China, in 1999 and 2000, respectively, and the Ph.D. degree in computer science from Shanghai Jiao Tong University, Shanghai, China, in 2005. 
He is currently a Full Professor with the Department of Computer Science and Engineering, Shanghai Jiao Tong University after he joined the University in 2009. 
He was a Research Fellow with the City University of Hong Kong, Hong Kong, from 2006 to 2009, a Visiting Scholar with Microsoft Research Asia, Beijing, China, in 2011, a Visiting Expert with the National Institute of Information and Communications Technology, Tokyo, Japan in 2012. He is an ACM Professional Member. 
He was the Area Co-Chair of ACL 2017 on tagging, chunking, syntax and parsing and Senior Area Chair of ACL 2018 and 2019 on phonology, morphology, and word segmentation. 
His research interests include natural language processing and related machine learning, data mining, and artificial intelligence.
\end{IEEEbiography}

\end{document}